\documentclass{article}
\usepackage{spconf,amsmath,graphicx,array}
\usepackage{booktabs}
\usepackage{multirow}
\usepackage{url}
\usepackage{arydshln}

\title{A Multi-To-One Interview Paradigm for Efficient MLLM Evaluation}

\name{Ye Shen$^{1,2}$, Junying Wang$^{2,3}$, Farong Wen$^{1,2}$, Yijin Guo$^{1,2}$, Qi Jia$^{2}$, Zicheng Zhang$^{\dagger2}$, Guangtao Zhai$^{\dagger1,2}$}
\address{$^1$Shanghai Jiao Tong University, 
$^2$Shanghai AI Laboratory, 
$^3$Fudan University\\
$^\dagger$Corresponding author. }

\begin{document}
\topmargin=0mm

\maketitle
 
\begin{figure*}[!ht]
\centering
\includegraphics[width=0.9\textwidth]{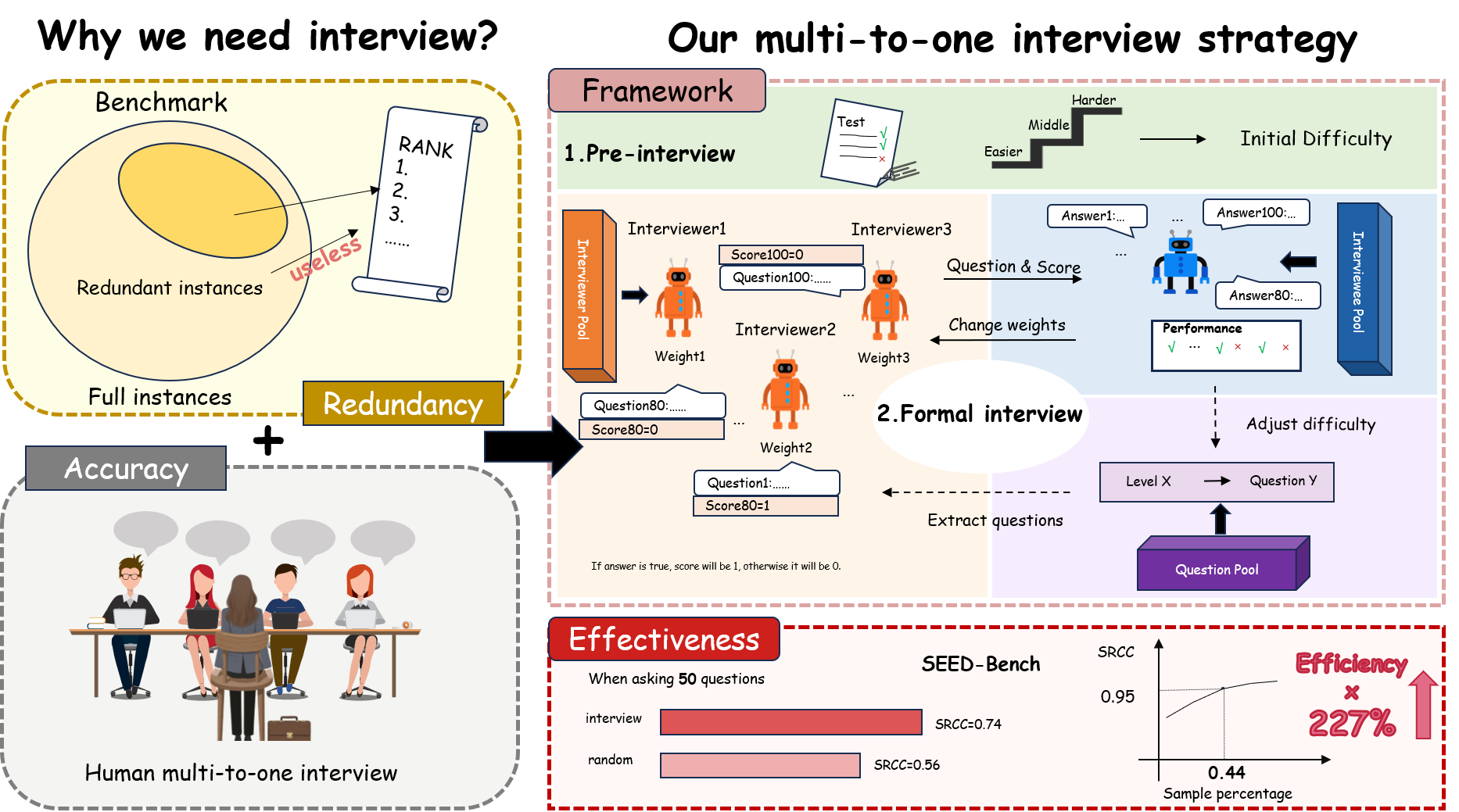}

\caption{Using full instances in the benchmark is less efficient for evaluation. Considering the efficiency of human multi-to-one interviews, we propose a multi-to-one interview paradigm in evaluating the capabilities of models and demonstrate its effectiveness, accuracy, and fairness.}
\label{fig:res}

\end{figure*}

\begin{abstract}

The rapid progress of Multi-Modal Large Language Models (MLLMs) has spurred the creation of numerous benchmarks. However, conventional full-coverage Question–Answering evaluations suffer from high redundancy and low efficiency. Inspired by human interview processes, we propose a \textbf{multi-to-one interview paradigm} for efficient MLLM evaluation. Our framework consists of (i) a two-stage interview strategy with pre-interview and formal interview phases, (ii) dynamic adjustment of interviewer weights to ensure fairness, and (iii) an adaptive mechanism for question difficulty-level chosen. Experiments on different benchmarks show that the proposed paradigm achieves significantly higher correlation with full-coverage results than random sampling, with improvements of up to 17.6\% in PLCC and 16.7\% in SRCC, while reducing the number of required questions. These findings demonstrate that the proposed paradigm provides a reliable and efficient alternative for large-scale MLLM benchmarking.

\end{abstract}
\begin{keywords}
MLLM Evaluation, Multi-To-One Interview
\end{keywords}

\section{Introduction}
\label{sec:intro}

Multi-Modal Large Language Models (MLLMs) have achieved remarkable performance across diverse tasks involving images, videos, audio, and 3D content~\cite{zhang2025lmmsurvey}. As these models advance rapidly, \textbf{reliable and efficient evaluation} has become a central research challenge, giving rise to a wide range of benchmarks~\cite{wang2025ever}. However, conventional full-coverage Question–Answering (Q\&A) evaluations suffer from \textbf{severe redundancy}: many instances are highly similar, contributing little new information to model assessment~\cite{zhang2025redundancy}. That is to say, fewer instances are sufficient for reliable ranking than for exhaustive evaluation, thus motivating the development of \textbf{more efficient paradigms for MLLM evaluation}.

Inspired by real-world hiring practices, where \textbf{multi-to-one interviews} enable evaluators to quickly gauge a candidate’s ability with a small number of well-chosen questions, we introduce a novel interview paradigm to address the inefficiency of full-coverage Q\&A testing. Our approach consists of three key components: (i) a \textbf{two-stage interview strategy}, including a lightweight pre-interview for initial difficulty calibration and a formal interview for comprehensive capability assessment; (ii) \textbf{dynamic adjustment of interviewer weights}, allowing diverse models to evaluate the interviewee more fairly and thoroughly; and (iii) an \textbf{adaptive difficulty mechanism} that updates subsequent questions based on both the current round’s difficulty and the interviewee’s performance, ensuring broad coverage of capability levels. Together, these components enable \textbf{comprehensive, accurate, fair, and efficient} evaluation of MLLMs.

Our main contributions are as follows:
 \begin{itemize}
     \item We propose a \textbf{multi-to-one interview paradigm} for MLLM evaluation, incorporating (i) a two-stage interview strategy, (ii) dynamic interviewer weight adjustment, and (iii) an adaptive difficulty mechanism.
    \item We demonstrate that this paradigm provides \textbf{reliable, fair, and efficient reflections} of MLLM capabilities, capturing both overall performance and difficulty-aware distributions.
    \item Extensive experiments on MMT-Bench, ScienceQA, and SEED-Bench show that our approach consistently outperforms random sampling and achieves up to \textbf{17.6\% PLCC and 16.7\% SRCC improvements} over full-coverage Q\&A testing with fewer questions.
\end{itemize}

\section{Related Work}
\label{sec:pagestyle}

\subsection{Multi-Model Large Language Models}

Early Large Language Models (LLMs) exhibit excellent text processing capabilities~\cite{fu2024mmesurveycomprehensivesurveyevaluation}. GPT-2~\cite{radford2019language} demonstrates an astonishing ability to converse with humans. Building upon LLMs, MLLMs are further equipped with capabilities of processing multimodal information ~\cite{fu2024mmesurveycomprehensivesurveyevaluation}. CLIP-ViT~\cite{radford2021learning} lays foundations for visual-textual alignment. Subsequently, models such as Claude-4-Sonnet~\cite{anthropic_claude4_2025}, GPT-4o~\cite{openai2024hello}, Grok-4~\cite{xai_grok4_2025}, and Gemini-2.5-Pro~\cite{team2023gemini} are constantly improving capabilities, excelling at numerous multimodel tasks, even complex tasks in real-world scenarios. However, how to evaluate the capabilities of models \textbf{more accurately, comprehensively, fairly, and realistically} has become an important issue. 
\subsection{MLLM Benchmarks}

The rapid evolution of MLLMs has driven benchmarks into a multi-dimensional capability comprehensive assessment~\cite{zhangaibench}. Early benchmarks such as VQA~\cite{antol2015vqa} assesses visual recognition ability, but fails to capture advanced reasoning capabilities. Consequently, researchers have begun to design comprehensive evaluation benchmarks tailored to MLLMs' characteristics. The dimensions covered by SEED-Bench~\cite{li2023seed} have increased significantly, while MMT-Bench~\cite{ying2024mmt} further expands the scope to complex tasks in actual application scenarios. Although benchmarks are constantly progressing, most of them use full-coverage Q\&A testing, \textbf{which often leads to redundancy and lack of adaptability, thus a more effective evaluation strategy is urgently required}.

\section{Interview Framework}

Totally, the well-designed interview framework combines (i) a two-stage interview strategy, (ii) dynamic adjustment of interviewer weights, and (iii) adaptive difficulty mechanism.
\subsection{The two-stage interview strategy}

The interview process includes an effective pre-interview to preliminarily divide difficulty levels and a formal interview to comprehensively evaluate the capabilities of models. Firstly, the pre-interview phase uses a written test to initially assess the model's capabilities. Specifically, middle-difficulty questions are randomly selected from the question pool to test the model. Then, corresponding accuracy is calculated to determine the initial difficulty of the formal interview. \begin{equation}
\text{Initial difficulty} = 
\begin{cases} 
middle + 1  & \text{if } acc > \beta, \\
middle & \text{if } acc = \beta, \\
middle - 1 & \text{otherwise},
\end{cases}
\end{equation} 
where $acc$ represents interviewee's test results, $\beta$ represents an adjustable accuracy threshold. Secondly, in each round of the formal interview, the selected interviewer is responsible for (i) extracting multiple classical questions from categories arranged in stack matching the current difficulty and (ii) judging whether the interviewee's answers are correct. Then according to the interviewee's performance, the difficulty of questions in each round will be adjusted. The whole interview ends once reaching the set number of questions.

\subsection{Dynamic adjustment of interviewer weights}

To improve the reliability of the interview paradigm, we introduce a strategy for dynamically adjusting interviewer weights. Specifically, multiple interviewers are selected from the interviewer pool, whose weights are equal at first. After a Q\&A, interviewers' weights are calculated based on all the interviewee's previous responses. The selected interviewer model that asks questions is determined by weights.
\begin{equation}
w_i^{t+1} = 
\begin{cases} 
(1-\alpha) \cdot w_i^t & \text{if } acc_i = 0 \text{ or } acc_i = 1, \\
(1+\alpha) \cdot w_i^t & \text{otherwise},
\end{cases}
\end{equation}
$w$ represents weight constrained between 0.5 and 2 to avoid extreme bias, $i$ represents the $i$-th interviewer, $t$ represents the $t$-th Q\&A, ${acc_i}$ represents performance of the interviewee under the inspection of the $i$-th interviewer, $\alpha$ is an adjustable weight threshold.

\subsection{Adaptive difficulty mechanism}

To improve the accuracy of the interview paradigm, we design a strategy for adaptively adjusting the difficulty of each round. In specific, after a round, accuracy is calculated to determine the difficulty of next round by formula (3). 
\begin{equation}
\text{L}^{r+1} = 
\begin{cases} 
\text{L}^{r} + 1 & \text{if } acc^{r} > \beta, \\
\text{L}^{r} & \text{if } acc^{r} =\beta, \\
\text{L}^{r} - 1 & \text{if } acc^{r} < \beta,
\end{cases}
\end{equation}
where L represents difficulty level, $r$ represents the $r$-th round, $acc^r$ represents the overall performance of the interviewee at the $r$-th round's difficulty level, $\beta$ represents an adjustable accuracy threshold. To prevent the interview from falling into local difficulty shock, if comes the situation that the difficulty change between ${Level}^{r-1}$ and ${Level}^{r}$ repeats three times, formula (4) is utilized to adjust the difficulty level appropriately.
\begin{equation}
\text{L}^{r+1} = 
\begin{cases} 
\text{L}^{r-1} + 3 & \text{if } \text{L}^{r-1}>n, \\
\text{L}^{r-1} -3 & \text{if } otherwise,\\
\end{cases}
\end{equation}
where $n$ represents an adjustable level threshold. And when no questions are available at ${Level}^{r}$, we adjust the difficulty level according to formula (5).
\begin{equation}
\text{L}^{r+1} = 
\begin{cases} 
\text{L}^{r} + 1 & \text{if } acc^{r}>\beta, \\
\text{L}^{r} - 1 & \text{if } otherwise. \\
\end{cases}
\end{equation}
 
\section{EXPERIMENT}

\subsection{Prepare work}

\subsubsection{The related benchmarks}

 To construct specific benchmarks with difficulty attributes, 10 typical models are selected to examine the questions, which includes GPT-4o~\cite{openai2024hello}, Deepseek-VL~\cite{wu2024deepseek}, Qwen-2.5-VL~\cite{bai2025qwen2}, Gemini-1.5-pro~\cite{team2024gemini}, Grok-2~\cite{xai_grok2_beta_2024}, Kimi-VL~\cite{team2025kimi}, Phi-3~\cite{abdin2024phi}, Claude3-7~\cite{anthropic2024claude}, HunYuan~\cite{sun2024hunyuan} and InternVL3~\cite{chen2024internvl}. Subtract the number of models that answered correctly from 11 to obtain the difficulty of each question. If level is calculated as 11, it is treated as 10. Ultimately, questions in MMT-Bench~\cite{ying2024mmt}, ScienceQA~\cite{saikh2022scienceqa} and SEED-Bench~\cite{li2023seed} are successfully divided into 10 difficulty levels.

\subsubsection{The related models}

The interviewer and interviewee models used in the experiments are shown in Tab.1.
\begin{table}[htbp]
    \begin{tabular}{>{\raggedright\arraybackslash}m{0.3\linewidth}>{\raggedright\arraybackslash}m{0.6\linewidth}}
        \hline
        \textbf{Calling Method} & \textbf{Model} \\
        \hline
    API Call & GPT-4.1-Nano~\cite{achiam2023gpt}, GPT-4o-Mini~\cite{openai2024hello}, GPT-4o~\cite{openai2024hello}, Grok-2~\cite{xai_grok2_beta_2024}, Grok-4~\cite{xai_grok4_2025}, Claude-3.5-Sonnet~\cite{anthropic2024claude}, Claude-4-Sonnet~\cite{anthropic_claude4_2025}, Qwen-VL-Plus~\cite{bai2023qwen}, Gemini-2.0-Flash~\cite{team2023gemini}, Gemini-2.5-Pro~\cite{team2023gemini}, Gemini-2.5-Flash~\cite{team2023gemini}\\
        \hline
        Local Call & Phi-4-mini-Instruct~\cite{abdin2024phi}, Qwen2.5-VL-7b-Instruct~\cite{bai2025qwen2}, Qwen2.5-VL-72B-Instruct~\cite{bai2025qwen2}, Qwen2.5-VL-32B-Instruct~\cite{bai2025qwen2}, Internlm3-8B-Instruct~\cite{cai2024internlm2}, Internlm2.5-20B-Chat~\cite{cai2024internlm2}, Gemma-3-27B-It~\cite{team2025gemma}, Llama-3.2-11B-Vision-Instruct~\cite{grattafiori2024llama} \\
        \hline
    \end{tabular}
     \caption{MLLM interviewees and MLLM interviewers illustration. 11 advanced closed source MLLMs and 8 advanced open source MLLMs are utilized.}  
    \label{tab:MLLMs}
\end{table}

\subsection{Settings}

Each model in Tab. 1 is designated as an interviewee, with 3 randomly selected via API as interviewers. The pre-interview administers 3 random Level 5 questions. Formal interview rounds each include 3 questions. Interviewers start with weight 1.0. Parameters are set, including $\alpha=0.2$ to ensure the rationality of weight adjustment, $\beta=0.5$ to accurately measure model level, and $n=7$ to cover full difficulties. The set number is 200.

Accuracy across different difficulties are calculated to verify interviewees' capability distribution and set up a control group where we randomly select questions from benchmarks. The performance of MLLMs in full-coverage Q\&A testing serves as the ground truth. 3 widely adopted metrics is utilized for analysis, including Spearman's Rank Correlation Coefficient (SRCC)~\cite{spearman1961proof}, Pearson's Linear Correlation Coefficient (PLCC)~\cite{pearson1895vii}, and Kendall's Rank Correlation Coefficient (KRCC)~\cite{kendall1938new}.

\begin{table*}[!ht]
  \centering
  \renewcommand\arraystretch{1.15}
  \renewcommand\tabcolsep{6pt}
  \caption{A comparison between the multi-to-one interview paradigm and the random strategy, where Avg.improvement means average absolute increase (percentage points) of our paradigm over random sampling in each metric.}
  \resizebox{\linewidth}{!} {\begin{tabular}{c|rrr|rrr|rrr}
    \hline
    {\textbf{Benchmark}} & \multicolumn{3}{c|}{\textbf{MMT-Bench}} & \multicolumn{3}{c|}{\textbf{ScienceQA}} & \multicolumn{3}{c}{\textbf{SEED-Bench}} \\
    \hline
    {\textbf{Number}}& \textbf{SRCC} & \textbf{PLCC} & \textbf{KRCC} & \textbf{SRCC} & \textbf{PLCC} & \textbf{KRCC} & \textbf{SRCC} & \textbf{PLCC} & \textbf{KRCC} \\
    \hline
    \multicolumn{10}{l}{\textit{Random Strategy: Questions picked randomly}} \\
    \hdashline
    20 & 0.5573 & 0.6654 & 0.4279 & 0.3515 & 0.4205 & 0.2284 & 0.4277 & 0.4391 & 0.3090 \\
    30 & 0.5687 & 0.6705 & 0.4377 & 0.3668 & 0.4316 & 0.2433 & 0.4590 & 0.4876 & 0.3283\\
    50 & 0.6397 & 0.7021 & 0.4955 & 0.4688 & 0.6183 & 0.3512 & 0.5577 & 0.5894 & 0.4085 \\
    80 & 0.6496 & 0.6985 & 0.5373 & 0.5592 & 0.6376 & 0.4345 & 0.6129 & 0.6925 & 0.5091 \\
    100 & 0.6932 & 0.6470 & 0.5426 & 0.5984 & 0.6402 & 0.4556 & 0.6554 & 0.6907 & 0.4820 \\
    \hline
    \multicolumn{10}{l}{\textit{Multi-to-One Interview Paradigm (proposed): Questions picked during interview}} \\
    \hdashline
    20 & 0.6202 & 0.7275 & 0.4273 & 0.5911 & 0.7255 & 0.4366 & 0.5958 & 0.6780 & 0.4372 \\
    30 & 0.6424 & 0.7608 & 0.4557 & 0.6289 & 0.7225 & 0.4956 & 0.7122 & 0.6579 & 0.5494 \\
    50 & 0.6749 & 0.7957 & 0.5148 & 0.6348 & 0.7262 & 0.5118 & 0.7377 & 0.7110 & 0.5733 \\
    80 & 0.7091 & 0.8303 & 0.5385 & 0.6435 & 0.7213 & 0.4882 & 0.7491 & 0.7239 & 0.5852 \\
    100 & 0.7109 & 0.8308 & 0.5503 & 0.6547 & 0.7328 & 0.5013 & 0.7532 & 0.7251 & 0.5962  \\
    \hline
    {\textbf{Avg. improvement}}& {\textbf{4.98\%}} & {\textbf{11.23\%}} & {\textbf{0.91\%}} & {\textbf{16.17\%}} & {\textbf{17.60\%}} & {\textbf{14.41\%}} & {\textbf{16.71\%}} & {\textbf{11.93\%}} & {\textbf{14.09\%}} \\
    \hline
  \end{tabular}}
  \label{tab:benchmark}
\end{table*}

\begin{figure*}[!ht]

\begin{minipage}[b]{.33\linewidth}
  \centering
  \centerline{\includegraphics[width=6cm]{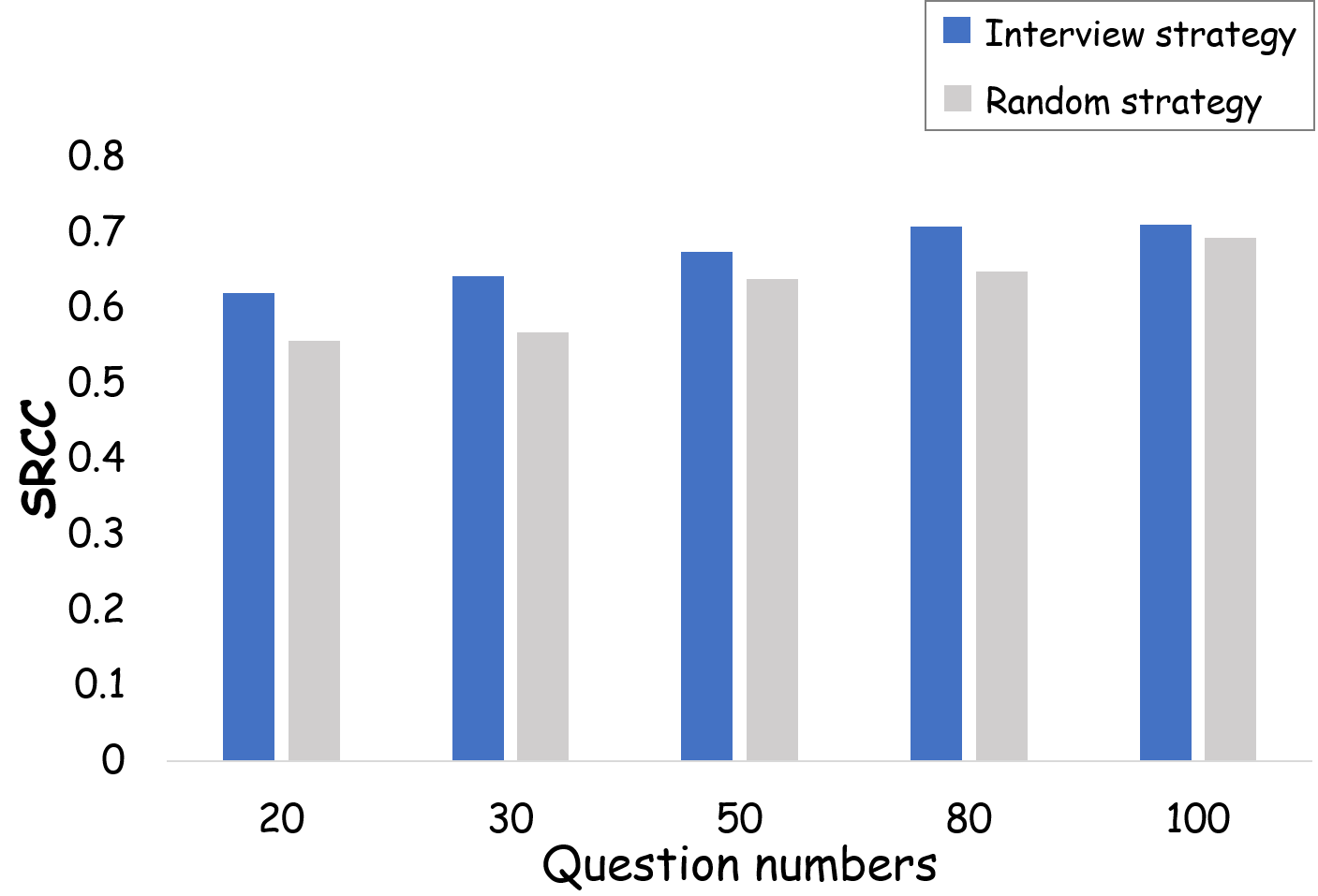}}
  \centerline{(a) MMT-Bench}\medskip
\end{minipage}
\begin{minipage}[b]{.33\linewidth}
  \centering
  \centerline{\includegraphics[width=6cm]{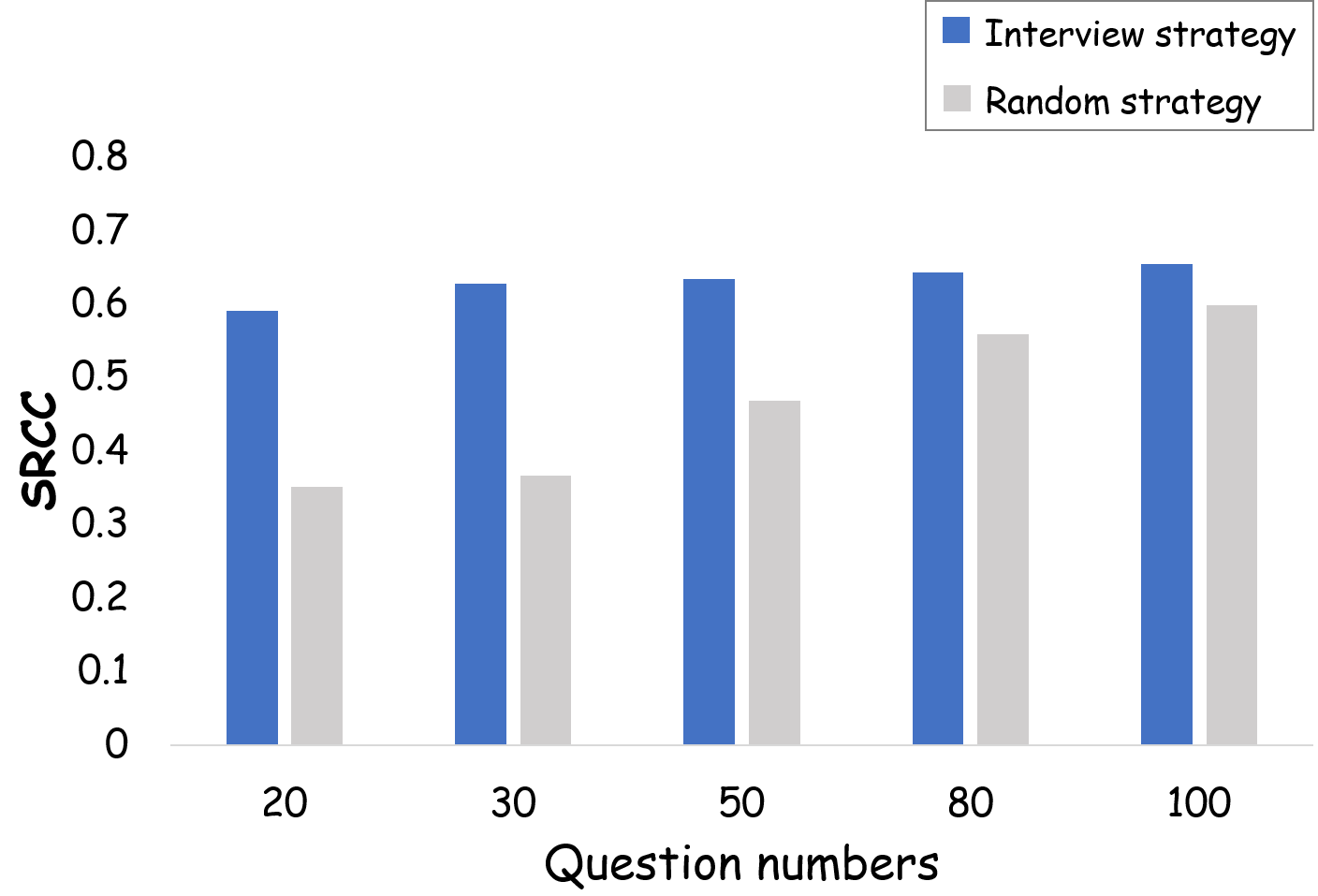}}
  \centerline{(b) ScienceQA}\medskip
\end{minipage}
\begin{minipage}[b]{0.33\linewidth}
  \centering
  \centerline{\includegraphics[width=6cm]{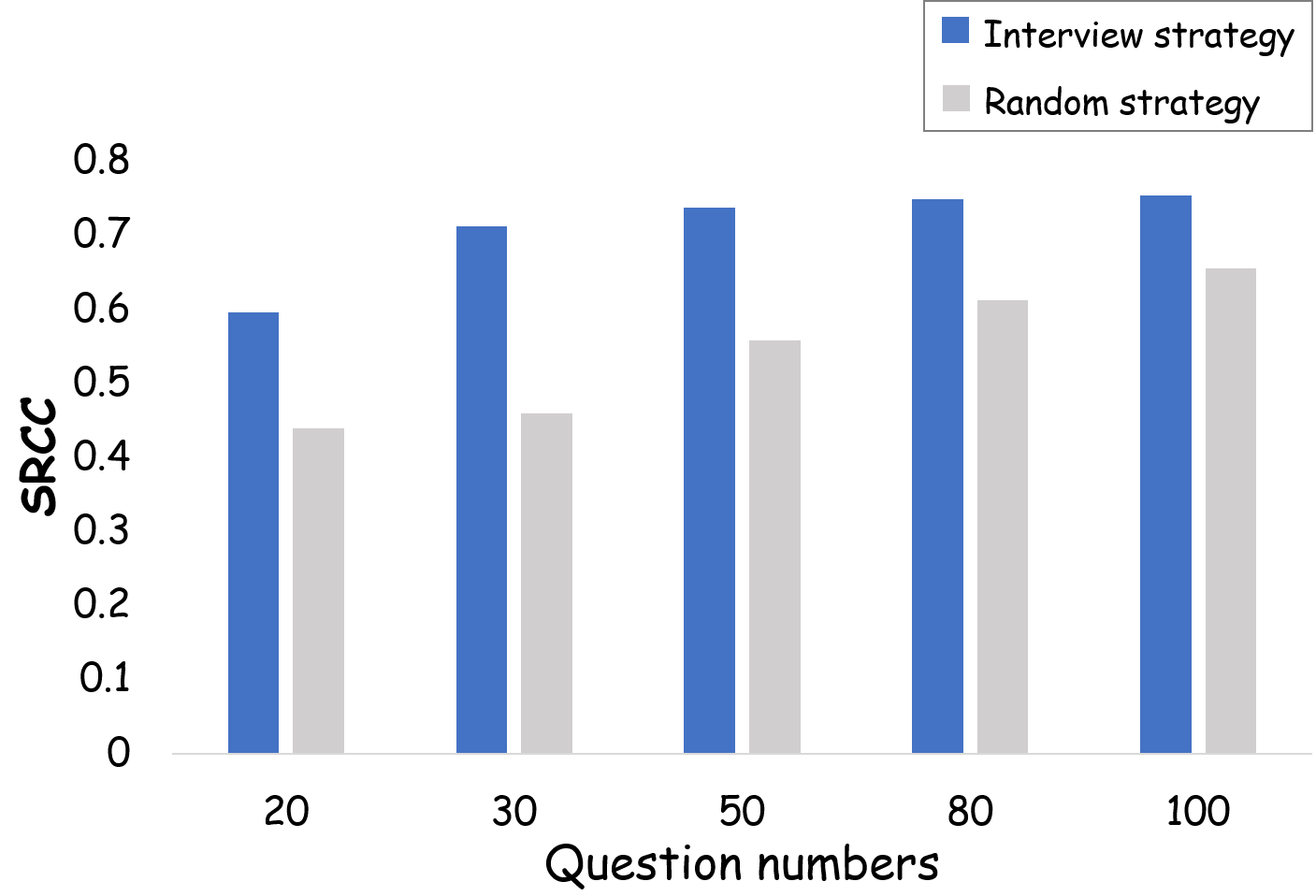}}
  \centerline{(c) SEED-Bench}\medskip
\end{minipage}
\caption{Overview of SRCC performance corresponding the question numbers on 3 Benchmarks clearly distinguishes the gap between the two methods.}
\label{fig:SRCC}
\end{figure*}

\subsection{Analysis}

The effectiveness of the paradigm is demonstrated across three benchmarks: MMT-Bench, ScienceQA, and SEED-Bench, compared with the random strategy.

\subsubsection{Accuracy of our paradigm}

According to the values revealed in Tab.2, the multi-to-one interview paradigm achieves higher SRCC, PLCC and KRCC values on most question settings than those of random sampling. For example, when asking 50 questions, the SRCC of proposed paradigm on SEED-Bench is 0.7377, while the random strategy is only 0.5577. This phenomenon, related to the huge number of questions in SEED-Bench, confirms that the interview paradigm can more accurately evaluate the capabilities of models with limited questions available.

\subsubsection{Efficiency of our paradigm}

The distinct gap of SRCC values illustrated in Fig.2 between two strategies validates the superior efficiency of our approach. When 30 questions are selected, the paradigm achieves an SRCC on ScienceQA of 0.6289, whereas the random strategy only reaches 0.3668, representing a 71.46\% improvement. This performance disparity can be attributed to ScienceQA questions being relatively difficult, but the interview paradigm can target the most informative questions early during the evaluation stage.

\section{CONCLUSION}

In summary, the multi-to-one interview paradigm enhances MLLM evaluation efficiency and reforms existing evaluation methods. However, there are still certain limitations including constraints of Q\&A Evaluation and relatively simple selection of interviewers. Hopefully, on one hand, it can be extended to automated benchmark construction; on the other hand, it holds significant potential in cross-lingual evaluation. 

\vfill\pagebreak

\bibliographystyle{IEEEbib}
\bibliography{main}

\end{document}